# COMBINING NAIVE BAYES AND DECISION TREE FOR ADAPTIVE INTRUSION DETECTION

Dewan Md. Farid[1], Nouria Harbi[1], and Mohammad Zahidur Rahman[2]

[1]ERIC Laboratory, University Lumière Lyon 2 - France
dewanfarid@gmail.com, nouria.harbi@univ-lyon2.fr
[2]Department of Computer Science and Engineering,
Jahangirnagar University, Bangladesh
rmzahid@juniv.edu

## ABSTRACT

*In this paper, a new learning algorithm for adaptive network intrusion detection using naive Bayesian classifier and decision tree is presented, which performs balance detections and keeps false positives at acceptable level for different types of network attacks, and eliminates redundant attributes as well as contradictory examples from training data that make the detection model complex. The proposed algorithm also addresses some difficulties of data mining such as handling continuous attribute, dealing with missing attribute values, and reducing noise in training data. Due to the large volumes of security audit data as well as the complex and dynamic properties of intrusion behaviours, several data mining-based intrusion detection techniques have been applied to network-based traffic data and host-based data in the last decades. However, there remain various issues needed to be examined towards current intrusion detection systems (IDS). We tested the performance of our proposed algorithm with existing learning algorithms by employing on the KDD99 benchmark intrusion detection dataset. The experimental results prove that the proposed algorithm achieved high detection rates (DR) and significant reduce false positives (FP) for different types of network intrusions using limited computational resources.*

## KEYWORDS

*Decision Tree, Detection Rate, False Positive, Naive Bayesian classifier, Network Intrusion Detection*

## 1. INTRODUCTION

An "Intrusion Detection System (IDS)" is a system for detecting intrusions that attempting to misuse the data or computing resources of a computer system. Mostly intrusions are the violation of information security policy. At first IDS was implemented for host-based that located in servers to examine the internal interfaces [1]-[3], but with the evolution of computer networks the focus gradually shifted toward network-based. Network intrusion detection system (NIDS) performs packet logging, real-time traffic analysis of IP network, and tries to discover if an intruder is attempting to break into the system [4]-[6]. Snort is an open source network intrusion detection and prevention system (NIDPS) developed by Sourcefire [7], [8]. Snort performs protocol analysis, content searching/matching, and commonly blocks a variety of intrusions such as buffer overflows, stealth port scans, web application attacks, SMB probes, and OS fingerprinting attempts. Normally, intruders in computer system are classified into two categories like internal and external intruders. Internal intruders are users in the network and have some authority, but seek to gain additional ability to take action without legitimate authorization. External intruders do not have any authorized access to the system that they attack. Two types of detection models: misuse and anomaly are commonly using by IDS. Misuse detection model performs simple pattern matching techniques to match an attack pattern corresponding to known attack patterns in the database and produces very low false positives (FP). Anomaly detection model identifies new attacks by analyzing the anomalous behaviors from normal behaviors [9], and achieves high detection rates (DR) for new attacks, but produces many false positives (FP). Anomaly based IDS generate rules by observing collected audit data that is the records of activities generated by the operating system. Currently adaptive intrusion detection aims to solve the problems of analyzing the huge volumes of audit data and realizing performance optimization of detection rules [10]-[14].

Detecting intrusions using data mining algorithms such as decision tree (DT), naïve Bayesian (NB) [15], neural network (NN), support vector machine (SVM) [16], k-nearest neighbors (KNN), fuzzy logic model [17], and genetic algorithm [18] have been widely used in the last decades. However, there exist various problems in current IDS such as low detection accuracy, unbalanced detection rates for different types of attacks, high false positives, redundancy of input attributes as well as examples in the training data. Another difficulty of current IDS is to detect intrusions in real time high-speed networks, because the high-speed networks require IDS to deal with large volumes of network data in a very short time. In this paper, based on a comprehensive analysis for the current research challenges in intrusion detection, a new learning algorithm for adaptive network intrusion detection using naive Bayesian classifier and decision tree is presented, which can handle the above mentioned challenging issues. This paper also addresses some difficulties of data mining such as handling continuous attribute, dealing with missing attribute values, and reducing noise in training data. The experimental results by using KDD99 benchmark intrusion detection dataset prove that the proposed algorithm has achieved both high detection rates (DR) and the significant reduction of false positives (FP) in comparison with existing methods.

The remainders of the paper are organized as follows. Section 2 presents the related overview of networking and intrusion detection. The basic problems of learning are discussed in Section 3, whereas our proposed algorithm is introduced in Section 4. Then, the experimental results are expressed in Section 5. Finally, our conclusions and future works are mentioned in Section 6.

## 2. NETWORKING AND INTRUSION DETECTION OVERVIEW

### 2.1. Networking Overview

In the networks, TCP/IP is widely used for network communications, which are composed of four layers: application layer, transport layer, network layer, and hardware layer that work together [19]. When data transfers across the networks, the data passes from the highest layer through intermediate layers to the lowest layer. The lowest layer sends the accumulated data to its destination through the physical network. Application layer sends and receives data for particular applications, such as DNS, HTTP, FTP, SMTP, and SNMP. It enables applications to transfer data between server and client, and passes application data to the transport layer. Transport layer is responsible for packaging data using TCP (Transmission control protocol) and UDP (user datagram protocol) so that it can be transmitted between hosts. Each TCP or UDP packet has a source port and a destination port number. Network layer or internet protocol (IP) layer is in charge of handling the addressing and routing of data received from the transport layer. After the network layer has encapsulated the transport layer data, the resulting logical units are referred to as packets. Each packet contains a header, which is composed of various fields. Commonly used network layer protocols are IPv4, IPv6, ICMP (internet control message protocol), and IGMP (internet group management protocol). It is also responsible for providing error and status information involving the addressing and routing of data. Hardware layer or data link layer tackles communications on the physical network components including cables, routers, switches, and network interface cards (NIC). The best known hardware layer protocol is Ethernet. Ethernet relies on the concept of a media access control (MAC) address, which is a unique six-byte value (such as 00-02-B4-DA-92-2C) permanently assigned to a particular NIC.

## 2.2. Intrusion Detection Overview

Intrusion detection is the process of monitoring and analyzing the events in computer systems or networks to discover the signals of possible incidents, which attempt to compromise the confidentiality, integrity, and availability of computer resources. In general, IDS use misuse-based and anomaly-based detection model for detecting intrusions. Misuse-based IDS are very effective for detecting known attacks but largely ineffective for detecting new attacks whose pattern has not stored in the database yet. It performs pattern matching to match an attack pattern corresponding to known attack patterns in the database. Anomaly-based IDS identify new attacks by analyzing anomalous behavior from normal behaviors. It has a relatively high detection rate for new attack, but produces many false positives. It uses profiles that are developed by monitoring the characteristics of typical activities over a period of time and then compares the characteristics of current activity to thresholds related to the profile. A network-based IDS (NIDS) monitor and analyze network traffics, and use multiple sensors for detecting intrusions from internal and external networks [20]-[22]. IDS analyze the information gathered by the sensors, and return a synthesis of the input of the sensors to system administrator or intrusion prevention system. System administrator carries out the prescriptions controlled by the IDS. Today, data mining has become an indispensable tool for analyzing the input of the sensors in IDS. Fig. 1 shows a scenario of IDS to protect server machine from internal and external network.

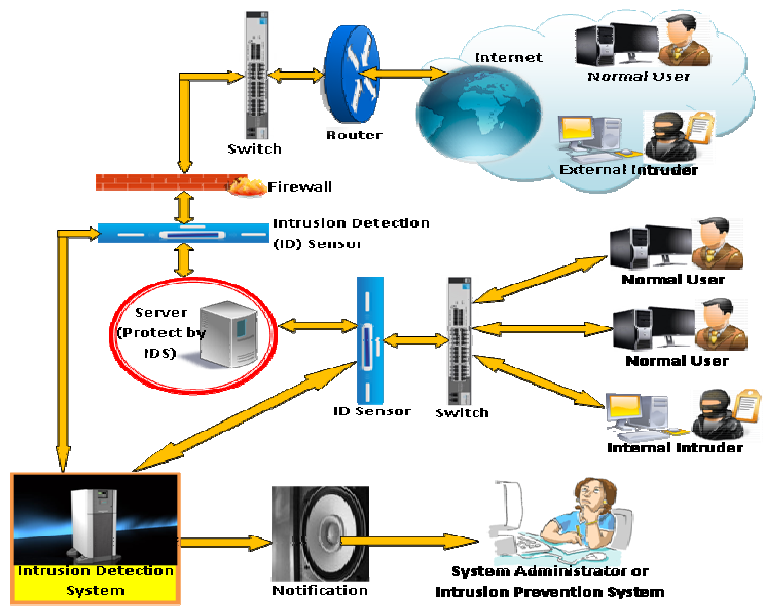

Figure 1. A typical motivating scenario of intrusion detection.

Ideally, IDS should have an attack detection rate (DR) of 100% along with false positive (FP) of 0%. Nevertheless, in practice this is really hard to achieve. The most important parameters involved in the performance estimation of IDS are shown in Table 1.

Table 1. Parameters for performance estimation of IDS.

| Parameters | Definition |
|---|---|
| True Positive (*TP*) or Detection Rate (*DR*) | Attack occur and alarm raised |
| False Positive (*FP*) | No attack but alarm raised |
| True Negative (*TN*) | No attack and no alarm |
| False Negative (*FN*) | Attack occur but no alarm |

The metrics such as precision, recall, overall, and false alarm have been used to measure the performance of the data mining algorithm on the minority class [23]-[26]. From Table 1, precision, recall, and overall may be defined as follows.

$$\text{Precision} = \frac{TP}{TP + FP} \tag{1}$$

$$\text{Recall} = \frac{TP}{TP + FN} \tag{2}$$

$$\text{Overall} = \frac{TP+TN}{TP+FP+FN+TN} \tag{3}$$

$$\text{False Alarm} = \frac{FP+FN}{TP+FP+FN+TN} \tag{4}$$

Detection rate (DR) and false positive (FP) are used to estimate the performance of IDS [27], which are given as bellow:

$$DR = \frac{Total\_detected\_attacks}{Total\_attacks} * 100 \tag{5}$$

$$FP = \frac{Total\_misclassifed\_process}{Total\_normal\_process} * 100 \tag{6}$$

### 2.3. Related Work

In 1980, the concept of intrusion detection began with Anderson's seminal paper [40]; he introduced a threat classification model that develops a security monitoring surveillance system based on detecting anomalies in user behavior. In 1986, Dr. Denning proposed several models for commercial IDS development based on statistics, Markov chains, time-series, etc [41]. In the early 1980's, Stanford Research Institute (SRI) developed an Intrusion Detection Expert System (IDES) that monitors user behavior and detects suspicious events [42]. In 1988, a statistical anomaly-based IDS was proposed by Haystack [43], which used both user and group-based anomaly detection strategies. In 1996, Forrest et al. proposed an analogy between the human immune system and intrusion detection that involved analyzing a program's system call sequences to build a normal profile [44]. In 2000, Valdes et al. [45] developed an anomaly based IDS that employed naïve Bayesian network to perform intrusion detecting on traffic bursts. In 2003, Kruegel et al. [46] proposed a multisensory fusion approach using Bayesian classifier for classification and suppression of false alarms that the outputs of different IDS sensors were aggregated to produce single alarm. In the same year, Shyu et al. [47] proposed an anomaly based intrusion detection scheme using principal components analysis (PCA), where PCA was applied to reduce the dimensionality of the audit data and arrive at a classifier that is a function of the principal components. In 2003, Yeung et al. [2] proposed an anomaly based intrusion detection using hidden Markov models that computes the sample likelihood of an observed sequence using the forward or backward algorithm for identifying anomalous. Lee et al. [48] proposed classification based anomaly detection using inductive rules to characterize sequences occurring in normal data. In 2000, Dickerson at al. [49] developed the Fuzzy Intrusion Recognition Engine (FIRE) using fuzzy logic that process the network data and generate fuzzy sets for every observed feature and then the fuzzy sets are used to detect network attacks. In 2003, Ramadas et al. [50] presented the anomalous network traffic detection with self organizing maps using DNS and HTTP services that the neurons are trained with normal network traffic then real time network data is fed to the trained neurons, if the distance of the incoming network traffic is more than a preset threshold then it raises an alarm.

## 3. BASIC PROBLEMS OF LEARNING

### 3.1. Handling Noise in Dataset

Noise in the dataset is considered to be one of the most challenging issues in data mining. This is because the performance of the learning algorithms depends on the quality of dataset. The main idea of dealing with noisy data at the learning time is to avoid over-fitting the dataset. Noise handling can be carried out at different stages of rule induction and interpretation. The followings exhibit some typical noises being existed in the dataset.

1) Missing attribute values: The simplest way for handling missing or unknown attribute value is to replace the missing attribute value with the most frequent attribute value in the dataset. Whereas, the most sophisticated way is to calculate the probability for attribute values and assign the probability value rather than the guessed value to each missing attribute value.

2) Contradictory examples: The same examples appear more than once in the dataset with different class labels. Contradictory examples confuse the learning algorithms, so these examples should be avoided or labeled correctly before learning.

3) Redundant examples: There often exist multiple copies of the same example in the dataset. Redundant examples are not a problem if they do not form contradictions, but this redundancy can change the decision trees produced by ID3 algorithm. For data mining, it's better to remove redundancy by keeping only a unique example in the dataset. By doing so, it not only saves the space of storage in dataset but also speeds significantly up the learning process.

4) Incomplete attribute problem: When the essential attributes of a problem are not used to describe in the dataset. Suppose to distinguish men from women based on the descriptions of a large group of people in terms of gender, height, weight, qualifications, and so on. The right attribute for men is "gender = male" and women is "gender = female". If we cannot catch the right attribute, then the classification model will be more complex and less accurate.

5) Misclassified examples: The examples in the dataset ware labeled with a wrong classification.

### 3.2. Dealing with Continuous Attribute

The goal of dealing with continuous attributes is to discretized the continuous attribute containing continuous values (i.e., real numbers or integers) into a number of intervals. The discretized intervals can be treated in a similar way to nominal values during learning and classification. It is very important for discretization of continuous attribute to find the right places to set up interval borders. The simplest technique is to place the interval borders of continuous attribute values between each adjacent pair of attribute values that are not classified into the same class. Suppose the pair of adjacent values on attribute $A_i$ are $A_{i1}$ and $A_{i2}$, "$A = (A_{i1}+A_{i2})/2$" can be taken as an interval border. The information gain technique adopted in ID3 algorithm is another very efficient technique to find the most informative border to split the values of the continuous attribute. The maximum information gain is always considered for a cut point (the midpoint) between the values taken by two attribute values of different classes. Each value of the formula "$A = (A_i + A_{i+1})/2$" where $i = 1,…,n-1$ is a possible cut point, if $A_i$ and $A_{i+1}$ have been taken by different class values in the dataset. The purpose of employing information gain is to check each of the possible cut points and find the best split point. In C4.5 algorithm, each of the possible cut points is not the midpoint between the two nearest values, rather than the greatest value in the entire dataset that does not exceed the midpoint. The naïve Bayesian classifier also uses for discretization of continuous attribute values by constructing a probability curve for each class in the dataset. When the curves for every class have been constructed, interval borders are placed on each of those points where the leading curves are different from its two sides. A few other methods such as equal distance division, grouping, k-nearest neighbors, and fuzzy borders are also applied to the continuous attributes for discretization.

### 3.3. Input Attribute Selection from Dataset

Effective input attribute selection from dataset before learning is very important, because irrelevant and redundant attributes of dataset may lead to complex classification model as well as reduce the classification accuracy [29]-[36]. In complex classification domains, input attributes of dataset may contain false correlations, which hamper the classification process. Some attributes in the dataset may be redundant, because the information they add is contained in other attributes. Also, some extra attributes can increase the computational time, and can have impact on the classification accuracy. Input attributes selection using data mining involves the selection of a subset of attributes $d$ from a total of $D$ original attributes of dataset, based on a given optimization principle that improves the performance of classifier. Finding a useful attribute subset is a form of search. Ideally, attribute selection methods search through the subsets of attributes, and try to find the best ones among the completing $2^N$ candidate subsets according to some evaluation function.

## 4. PROPOSED HYBRID ALGORITHM

### 4.1. Proposed Learning Algorithm

Given a training data $D = \{t_1,...,t_n\}$ where $t_i = \{t_{i1},...,t_{ih}\}$ and the training data $D$ contains the following attributes $\{A_1, A_2,...,A_n\}$ and each attribute $A_i$ contains the following attribute values $\{A_{i1}, A_{i2},...,A_{ih}\}$. The attribute values can be discrete or continuous. Also the training data $D$ contains a set of classes $C = \{C_1, C_2,...,C_m\}$. Each example in the training data $D$ has a particular class $C_j$. The algorithm first searches for the multiple copies of the same example in the training data $D$, if found then keeps only one unique example in the training data $D$ (suppose all attribute values of two examples are equal then the examples are similar). Then the algorithm discreties the continuous attributes in the training data $D$ by finding each adjacent pair of continuous attribute values that are not classified into the same class value for that continuous attribute. Next the algorithm calculates the prior $P(C_j)$ and conditional $P(A_{ij}|C_j)$ probabilities in the training data $D$. The prior probability $P(C_j)$ for each class is estimated by counting how often each class occurs in the training data $D$. For each attribute $A_i$ the number of occurrences of each attribute value $A_{ij}$ can be counted to determine $P(A_i)$. Similarly, the conditional probability $P(A_{ij}|C_j)$ for each attribute values $A_{ij}$ can be estimated by counting how often each attribute value occurs in the class in the training data $D$. Then the algorithm classifies all the examples in the training data $D$ with these prior $P(C_j)$ and conditional $P(A_{ij}|C_j)$ probabilities. For classifying the examples, the prior and conditional probabilities are used to make the prediction. This is done by combining the effects of the different attribute values from that example. Suppose the example $e_i$ has independent attribute values $\{A_{i1}, A_{i2},...,A_{ip}\}$, we know $P(A_{ik} | C_j)$, for each class $C_j$ and attribute $A_{ik}$. We then estimate $P(e_i | C_j)$ by

$$P(e_i | C_j) = P(C_j) \prod_{k=1 \to p} P(A_{ij} | C_j) \qquad (7)$$

To classify the example, the algorithm estimates the likelihood that $e_i$ is in each class. The probability that $e_i$ is in a class is the product of the conditional probabilities for each attribute value with prior probability for that class. The posterior probability $P(C_j | e_i)$ is then found for each class and the example classifies with the highest posterior probability for that example.

After classifying all the training examples, the class value for each example in training data $D$ updates with Maximum Likelihood (ML) of posterior probability $P(C_j|e_i)$.

$$C_j = C_i \rightarrow P_{ML}(C_j|e_i). \tag{8}$$

Then again the algorithm calculates the prior $P(C_j)$ and conditional $P(A_{ij}|C_j)$ probabilities using updated class values in the training data $D$, and again classifies all the examples of training data using these probabilities. If any of the training example is misclassified then the algorithm calculates the information gain for each attributes $\{A_1, A_2,…,A_n\}$ in the training data $D$.

$$Info(D) = -\sum_{j=1}^{m} \frac{freq(C_j,D)}{|D|} \log_2 \left( \frac{freq(C_j,D)}{|D|} \right) \tag{9}$$

$$Info(T) = \sum_{i=1}^{n} \frac{|T_i|}{|T|} \inf o(T_i) \tag{10}$$

*Information Gain* $(A_i) = Info(D) - Info(T)$ (11)

And chooses one of the best attributes $A_i$ among the attributes $\{A_1, A_2,…,A_n\}$ from the training data $D$ with highest information gain value, Then split the training data $D$ into sub-datasets $\{D_1, D_2,…,D_n\}$ depending on the chosen attribute values of $A_i$. After the algorithm estimates the prior and conditional probabilities for each sub-dataset $D_i = \{D_1, D_2,…,D_n\}$ and classifies the examples of each sub-dataset $D_i$ using their respective probabilities. If any example of any sub-dataset $D_i$ is misclassified then the algorithm calculates the information gain of attributes for that sub-dataset $D_i$, and chooses the best attribute $A_i$ with maximum information gain value from sub-dataset $D_i$, and split the sub-dataset $D_i$ into sub-sub-datasets $D_{ij}$. Then again calculates the prior and conditional probabilities for each sub-sub-dataset $D_{ij}$, and also classifies the examples of sub-sub-datasets using their respective probabilities. The algorithm will continue this process until all the examples of sub/sub-sub-datasets are correctly classified. When the algorithm correctly classifies all the examples then the prior and conditional probabilities for each datasets are preserved for future classification of unseen examples. The main procedure of proposed algorithm is described as follows.

**Algorithm**

Input: Training Data, $D$

Output: Adaptive Intrusion Detection Model, AIDM

Procedure:

Step 1: Search the multiple copies of same example in $D$, if found then keeps only one unique example in $D$.

Step 2: For each continuous attributes in $D$ find the each adjacent pair of continuous attribute values that are not classified into the same class value for that continuous attribute

Step 3: Calculate the prior probabilities $P(C_j)$ and conditional probabilities $P(A_{ij}|C_j)$ in $D$.

Step 4: Classify all the training examples using these prior and conditional probabilities, $P(e_i | C_j) = P(C_j) \prod_{k=1 \rightarrow p} P(A_{ij} | C_j)$.

Step 5: Update the class value for each example in $D$ with Maximum Likelihood (ML) of posterior probability, $P(C_j|e_i)$; $C_j = C_i \rightarrow P_{ML}(C_j|e_i)$.

Step 6: Recalculate the prior $P(C_j)$ and conditional $P(A_{ij}|C_j)$ probabilities using updated class values in $D$.

Step 7: Again classify all training examples in *D* using updated probability values.

Step 8: If any training examples in *D* is misclassified then calculate the information gain for each attributes $A_i = \{A_1, A_2,…,A_n\}$ in *D* using equation (11).

Step 9: Choose the best attribute $A_i$ from *D* with the maximum information gain value.

Step 10: Split dataset *D* into sub-datasets $\{D_1, D_2,…,D_n\}$ depending on the attribute values of $A_i$.

Step 11: Calculate the prior $P(C_j)$ and conditional $P(A_{ij}|C_j)$ probabilities of each sub-dataset $D_i$.

Step 12: Classify the examples of each sub-dataset $D_i$ with their respective prior and conditional probabilities.

Step 13: If any example of any sub-dataset $D_i$ is misclassified then again calculate the information gain of attributes for that sub-dataset $D_i$, and choose one of the best attribute $A_i$ with maximum information gain, then split the sub-dataset $D_i$ into sub-sub-datasets $D_{ij}$. Then again calculate the probabilities for each sub-sub-dataset $D_{ij}$. Also classify the examples in sub-sub-datasets using their respective probabilities.

Step 14: Continue this process until all the examples are correctly classified.

Step 15: Preserved all the prior and conditional probabilities for each dataset for future classification of unseen examples.

## 5. EXPERIMENTAL ANALYSIS

### 5.1. KDD Cup 1999 Dataset

The KDD cup 1999 dataset was used in the 3[rd] International Knowledge Discovery and Data Mining Tools Competition for building a network intrusion detector, a predictive model capable of distinguishing between intrusions and normal connections [37]. In 1998, DARPA intrusion detection evaluation program, a simulated environment was set up to acquire raw TCP/IP dump data for a local-area network (LAN) by the MIT Lincoln Lab to compare the performance of various intrusion detection methods. It was operated like a real environment, but being blasted with multiple intrusion attacks and received much attention in the research community of adaptive intrusion detection. The KDD99 dataset contest uses a version of DARPA98 dataset. In KDD99 dataset, each example represents attribute values of a class in the network data flow, and each class is labelled either normal or attack. The classes in KDD99 dataset can be categorized into five main classes (one normal class and four main intrusion classes: probe, DOS, U2R, and R2L).

1) Normal connections are generated by simulated daily user behaviour such as downloading files, visiting web pages.

2) Denial of Service (DoS) attack causes the computing power or memory of a victim machine too busy or too full to handle legitimate requests. DoS attacks are classified based on the services that an attacker renders unavailable to legitimate users like apache2, land, mail bomb, back, etc.

3) Remote to User (R2L) is an attack that a remote user gains access of a local user/account by sending packets to a machine over a network communication, which include send-mail, and Xlock.

4) User to Root (U2R) is an attack that an intruder begins with the access of a normal user account and then becomes a root-user by exploiting various vulnerabilities of the system. Most common exploits of U2R attacks are regular buffer-overflows, load-module, Fd-format, and Ffb-config.

5) Probing (Probe) is an attack that scans a network to gather information or find known vulnerabilities. An intruder with a map of machines and services that are available on a network can use the information to look for exploits.

In KDD99 dataset these four attacks (DoS, U2R, R2L, and probe) are divided into 22 different attacks that tabulated in Table 2.

Table 2. Different Types of attacks in KDD99 Dataset.

| 4 Main Attack Classes | 22 Attacks Classes |
|---|---|
| Denial of Service (DoS) | back, land, neptune, pod, smurt, teardrop |
| Remote to User (R2L) | ftp_write, guess_passwd, imap, multihop, phf, spy, warezclient, warezmaster |
| User to Root (U2R) | buffer_overflow, perl, loadmodule, rootkit |
| Probing | ipsweep, nmap, portsweep, satan |

There are total 41 input attributes in KDD99 dataset for each network connection that have either discrete or continuous values and divided into three groups. The first group of attributes is the basic features of network connection, which include the duration, prototype, service, number of bytes from source IP addresses or from destination IP addresses, and some flags in TCP connections. The second group of attributes in KDD99 is composed of the content features of network connections and the third group is composed of the statistical features that are computed either by a time window or a window of certain kind of connections. The list of the input attributes in KDD99 dataset for each network connections is shown in the Table 3.

Table 3. Input attributes in KDD99 Dataset.

| No | Input Attribute | Type | No | Input Attribute | Type |
|---|---|---|---|---|---|
| 1 | Duration | Con. | 22 | is_guest_login | Dis. |
| 2 | protocol_type | Dis. | 23 | Count | Con. |
| 3 | Service | Dis. | 24 | srv_count | Con. |
| 4 | Flag | Dis. | 25 | serror_rate | Con. |
| 5 | src_bytes | Con. | 26 | srv_serror_rate | Con. |
| 6 | dst_bytes | Con. | 27 | rerror_rate | Con. |
| 7 | Land | Dis. | 28 | srv_rerror_rate | Con. |
| 8 | wrong_fragment | Con. | 29 | same_srv_rate | Con. |
| 9 | Urgent | Con. | 30 | diff_srv_rate | Con. |
| 10 | Hot | Con. | 31 | srv_diff_host_rate | Con. |
| 11 | num_failed_logins | Con. | 32 | dst_host_count | Con. |
| 12 | logged_in | Dis. | 33 | dst_host_srv_count | Con. |
| 13 | num_compromised | Con. | 34 | dst_host_same_srv_rate | Con. |
| 14 | root_shell | Con. | 35 | dst_host_diff_srv_rate | Con. |
| 15 | su_attempted | Con. | 36 | dst_host_same_src_port_rate | Con. |
| 16 | num_root | Con. | 37 | dst_host_srv_diff_host_rate | Con. |
| 17 | num_file_creations | Con. | 38 | dst_host_serror_rate | Con. |
| 18 | num_shells | Con. | 39 | dst_host_srv_serror_rate | Con. |
| 19 | num_access_files | Con. | 40 | dst_host_rerror_rate | Con. |
| 20 | num_outbound_cmds | Con. | 41 | dst_host_srv_rerror_rate | Con. |
| 21 | is_host_login | Dis. | - | - | - |

Table 4 shows the number of examples in 10% training and testing data of KDD99 dataset.

Table 4. Number of training and testing examples in KDD99 Dataset.

| Attack Types | Training Examples | Testing Examples |
|---|---|---|
| Normal | 97277 | 60592 |
| Denial of Service | 391458 | 237594 |
| Remote to User | 1126 | 8606 |
| User to Root | 52 | 70 |
| Probing | 4107 | 4166 |
| Total Examples | 494020 | 311028 |

### 5.2. Experimental Analysis

In order to evaluate the performance of proposed algorithm for network intrusion detection, we performed 5-class classification using KDD99 intrusion detection benchmark dataset. All experiments were performed using an Intel Core 2 Duo Processor 2.0 GHz processor (2 MB Cache, 800 MHz FSB) with 1 GB of RAM. The results of the comparison of proposed algorithm with ID3 and with naive Bayesian classifier are in Table 5 and Table 6.

Table 5. Comparison of the results using 41 attributes.

| Method | Normal | Probe | DOS | U2R | R2L |
|---|---|---|---|---|---|
| Proposed Algorithm (DR %) | 99.72 | 99.25 | 99.75 | 99.20 | 99.26 |
| Proposed Algorithm (FP %) | 0.06 | 0.39 | 0.04 | 0.11 | 6.81 |
| Naïve Bayesian (DR %) | 99.27 | 99.11 | 99.69 | 64.00 | 99.11 |
| Naïve Bayesian (FP %) | 0.08 | 0.45 | 0.04 | 0.14 | 8.02 |
| ID3 (DR %) | 99.63 | 97.85 | 99.51 | 49.21 | 92.75 |
| ID3 (FP %) | 0.10 | 0.55 | 0.04 | 0.14 | 10.03 |

Table 6. Comparison of the results using 19 attributes.

| Method | Normal | Probe | DOS | U2R | R2L |
|---|---|---|---|---|---|
| Proposed Algorithm (DR %) | 99.84 | 99.75 | 99.76 | 99.47 | 99.35 |
| Proposed Algorithm (FP %) | 0.05 | 0.28 | 0.03 | 0.10 | 6.22 |
| Naïve Bayesian (DR %) | 99.65 | 99.35 | 99.71 | 64.84 | 99.15 |
| Naïve Bayesian (FP %) | 0.05 | 0.32 | 0.04 | 0.12 | 6.87 |
| ID3 (DR %) | 99.71 | 98.22 | 99.63 | 86.11 | 97.79 |
| ID3 (FP %) | 0.06 | 0.51 | 0.04 | 0.12 | 7.34 |

We tested the performance of proposed algorithm using the reduced dataset of 12 attributes and 17 attributes in KDD99 dataset, which increase the detection rate for intrusion classes that are summarized in Table 7.

Table 7. Performance of proposed algorithm using reduced dataset.

| Class Value | 12 Attributes | 17 Attributes |
|---|---|---|
| Normal | 99.98 | 99.95 |
| Probe | 99.92 | 99.93 |
| DoS | 99.99 | 99.97 |
| U2R | 99,38 | 99.46 |
| R2L | 99.55 | 99.69 |

## 6. CONCLUSIONS AND FUTURE WORKS

This paper introduced a new hybrid learning algorithm for adaptive network intrusion detection using naive Bayesian classifier and ID3 algorithm, which analyzes the large volume of network data and considers the complex properties of attack behaviours to improve the performance of detection speed and detection accuracy. In this paper we have concentrated on the development of the performance of naïve Bayesian classifier and ID3 algorithm. It has been successfully tested that this hybrid algorithm minimized false positives, as well as maximize balance detection rates on the 5 classes of KDD99 benchmark dataset. The attacks of KDD99 dataset detected with 99% accuracy using proposed algorithm. The future work focus on improving the false positives of remote to user (R2L) attack and apply this detection model into real world IDS.

## ACKNOWLEDGEMENTS

Support for this research received from ERIC Laboratory, University Lumière Lyon 2 – France, and Department of Computer Science and Engineering, Jahangirnagar University, Bangladesh.

## REFERENCES

[1] Jackson, T., Levine, J., Grizzard, J., Owen, and H., "An investigation of a compromised host on a honeynet being used to increase the security of a large enterprise network," IEEE workshop on Information Assurance and Security, IEEE, 2004.

[2] D. Y. Yeung, and Y. X. Ding, "Host-based intrusion detection using dynamic and static behavioral models," *Pattern Recognition*, 36, 2003, pp. 229-243.

[3] X. Xu, and T. Xie, "A reinforcement learning approach for host-based intrusion detection using sequences of system calls," In Proc. of International Conference on Intelligent Computing, Lecture Notes in Computer Science, LNCS 3644, 2005, pp. 995-1003.

[4] Krasser, S., Grizzard, J., Owen, H., and Levine. J., "The use of honeynets to increase computer network security and user awareness," Journal of Security Education, vol. 1, 2005, pp. 23-37.

[5] Shon T., Seo J., and Moon J., "SVM approach with a genetic algorithm for network intrusion detection," in Proc. of 20[th] International Symposium on Computer and Information Sciences (ISCIS 2005), Berlin: Springer-Verlag, 2005, pp. 224-233.

[6] X. Xu, X.N. Wang, "Adaptive network intrusion detection method based on PCA and support vector machines," Lecture Notes in Artificial Intelligence (ADMA 2005), LNAI 3584, 2005, pp. 696-703.

[7] Martin Roesch, "SNORT: The open source network intrusion system," Official web page of Snort at http://www.snort.org/

[8] L. C. Wuu, C. H. Hung, and S. F. Chen, "Building intrusion pattern miner for sonrt network intrusion detection system," Journal of Systems and Software, vol. 80, Issue 10, 2007, pp. 1699-1715.

[9] Lazarevic, A., Ertoz, L., Kumar, V., Ozgur,. A., Srivastava, and J., "A comparative study of anomaly detection schemes in network intrusion detection," In Proc. of the SIAM Conference on Data Mining, 2003.

[10] Sebastiaan Tesink, "Improving intrusion detection system through machine learning," Technical Report, Series no. 07-02, ILK Research Group, Tilburg University, March, 2007.

[11] Barbara, Daniel, Couto, Julia, Jajodia, Sushil, Popyack, Leonard, Wu, and Ningning, "ADAM: Detecting intrusion by data mining," IEEE Workshop on Information Assurance and Security, West Point, New York, June 5-6, 2001.

[12]  Lee W., "A data mining and CIDF based approach for detecting novel and distributed intrusions," Recent Advances in Intrusion Detection, 3rd International Workshop, RAID 2000, Toulouse, France, October 2-4, 2000, Proc. Lecture Notes in Computer Science 1907 Springer, 2000, pp. 49-65.

[13]  Lee W., Stolfo S., and Mok K., "Adaptive Intrusion Detection: A Data Mining Approach," Artificial Intelligence Review, 14(6), December 2000, pp. 533-567.

[14]  Stolfo J., Fan W., Lee W., Prodromidis A., and Chan P.K., "Cost-based modeling and evaluation for data mining with application to fraud and intrusion detection," DARPA Information Survivability Conference, 2000.

[15]  N.B. Amor, S. Benferhat, and Z. Elouedi, "Naïve Bayes vs. decision trees in intrusion detection systems," In Proc. of 2004 ACM Symposium on Applied Computing, 2004, pp. 420-424.

[16]  Mukkamala S., Janoski G., and Sung A.H., "Intrusion detection using neural networks and support vector machines," In Proc. of the IEEE International Joint Conference on Neural Networks, 2002, pp.1702-1707.

[17]  J. Luo, and S.M. Bridges, "Mining fuzzy association rules and fuzzy frequency episodes for intrusion detection," International Journal of Intelligent Systems, John Wiley & Sons, vol. 15, no. 8, 2000, pp. 687-703.

[18]  YU Yan, and Huang Hao, "An ensemble approach to intrusion detection based on improved multi-objective genetic algorithm," Journal of Software, vol. 18, no. 6, June 2007, pp. 1369-1378.

[19]  Karen Scarfone, and Peter Mell, "Guide to intrusion detection and prevention systems (IDPS)," National Institute of Standards and Technology, Gaithersburg, MD 20899-8930, NIST Special Publication 800-94, February, 2007.

[20]  Wasniowski R., "Multi-sensor agent-based intrusion detection system," In Proc. of the 2nd Annual Conference on Information Security, Kennesaw, Georgia, 2005, pp. 100-103.

[21]  Diego Zamboni, "Using internal sensors for computer intrusion detection," Ph.D. dissertation, Purdue University, 2001.

[22]  T. Bass, "Intrusion detection systems and multi-sensor data fusion," Communications of the ACM, 43(4), 2000, pp. 99-105.

[23]  F. Provost, and T. Fawcett, "Robust classification for imprecise environment," Machine Learning, vol. 42/3, 2001, pp. 203-231.

[24]  M. Joshi, V. Kumar, and R. Agarwal, "Evaluating boosting algorithms to classify rare classes: Comparison and Improvements," In Proc. of the 1st IEEE conference on Data Mining, San Jose, CA, 2001.

[25]  M. Joshi, R. Agarwal, and V. Kumar, "Predicting rare classes: can boosting make any weak learner stronger?," in Proc. of 8th ACM Conference ACM SIGKDD International Conference on Knowledge Discovery and Data Mining, Edmonton, Canada, 2002.

[26]  G. Helmer, J.S.K. Wong, V. Honavar, and L. Miller, "Automated discovery of concise predictive rules for intrusion detection," Journal of Systems and Software, vol. 60, no. 3, 2002, pp. 165-175.

[27]  R.C. Chen, and S.P. Chen, "Intrusion detection using a hybrid support vector machine based on entropy and TF-IDF," International Journal of Innovative Computing, Information, and Control (IJICIC), vol. 4, no. 2, 2008, pp. 413-424.

[28]  R. Agarwal, and M.V. Joshi, "PNrule: a new framework for learning classifier models in data mining (a case-study in network intrusion detection)," Proceedings of 1st SIAM Conference on Data Mining, 2001.

[29]     Radtke PVW, Sabourin R, and Wong T, "Intelligent feature extraction for ensemble of classifiers," In Proc. of 8th International Conference on Document Analysis and Recognition (ICDAR 2005), Seoul: IEEE Computer Society, 2005, pp. 866-870.

[30]     R. Rifkin, A. Klautau, "In defense of one-vs-all classification," Journal of Machine Learning Research, 5, 2004, pp. 143-151.

[31]     Chebrolu S., Abraham A., and Thomas J.P., "Feature deduction and ensemble design of intrusion detection systems," Computer & Security, 24(4), 2004, pp. 295-307.

[32]     Tsymbal A., Puuronen S., Patterson D.W., "Ensemble feature selection with the simple Bayesian classification," Information Fusion, 4(2), 2003, pp. 87-100.

[33]     Sung A.H., and Mukkamala S., "Identifying important features for intrusion detection using support vector machines and neural networks," In Proc. of International Symposium on Applications and the Internet (SAINT 2003), 2003, pp. 209-217.

[34]     Oliveira LS, Sabourin R, Bortolozzi RF, and Suen CY, "Feature selection using multi-objective genetic algorithms for handwritten digit recognition," In Proc. of 16th International Conference on Pattern Recognition (ICPR 2002), Quebec: IEEE Computer Society, 2002, pp. 568-571.

[35]     Mukkamala S., and Sung A.H., "Identifying key features for intrusion detection using neural networks," In Proc. of the ICCC International Conference on Computer Communications, 2002.

[36]     Lee WK, and Stolfo SJ, "A framework for constructing features and models for intrusion detection systems," ACM Transactions on Information and System Security, 3(4), 2000, pp. 227-261.

[37]     The KDD Archive. KDD99 cup dataset, 1999. http://kdd.ics.uci.edu/databases/kddcup99/kddcup99.html

[38]     M. Tavallaee, E. Bagheri, W. Lu, and A. Ghorbani, "A Detailed Analysis of the KDD CUP 99 Data Set," Submitted to Second IEEE Symposium on Computational Intelligence for Security and Defense Applications (CISDA) 2009.

[39]     J. McHugh, "Testing intrusion detection systems: a critique of the 1998 and 1999 darpa intrusion detection system evaluations as performed by Lincoln laboratory," ACM Transactions on Information and System Security, vol. 3. Np. 4, 2000, pp. 262-294.

[40]     James P. Anderson, "Computer security threat monitoring and surveillance," Technical Report 98-17, James P. Anderson Co., Fort Washington, Pennsylvania, USA, April 1980.

[41]     Dorothy E. Denning, "An intrusion detection model," IEEE Transaction on Software Engineering, SE-13(2), 1987, pp. 222-232.

[42]     Dorothy E. Denning, and P.G. Neumann "Requirement and model for IDES- A real-time intrusion detection system," Computer Science Laboratory, SRI International, Menlo Park, CA 94025-3493, Technical Report # 83F83-01-00, 1985.

[43]     S.E. Smaha, and Haystack, "An intrusion detection system," in Proc. of the IEEE Fourth Aerospace Computer Security Applications Conference, Orlando, FL, 1988, pp. 37-44.

[44]     S. Forrest, S.A. Hofmeyr, A. Somayaji, T.A. Longstaff, "A sense of self for Unix processes," in Proc. of the IEEE Symposium on Research in Security and Privacy, Oakland, CA, USA, 1996, pp. 120-128.

[45]     A. Valdes, K. Skinner, "Adaptive model-based monitoring for cyber attack detection," in Recent Advances in Intrusion Detection Toulouse, France, 2000, pp. 80-92.

[46]     C. Kruegel, D. Mutz, W. Robertson, F. Valeur, "Bayesian event classification for intrusion detection," in Proc. of the 19th Annual Computer Security Applications Conference, Las Vegas, NV, 2003.

[47]  M.L. Shyu, S.C. Chen, K. Sarinnapakorn, L. Chang, "A novel anomaly detection scheme based on principal component classifier," in Proc. of the IEEE Foundations and New Directions of Data Mining Workshop, Melbourne, FL, USA, 2003, pp. 172-179.

[48]  W. Lee, S.J. Stolfo, "Data mining approaches for intrusion detection," In Proc. of the 7th USENIX Security Symposium (SECURITY-98), Berkeley, CA, USA, 1998, pp. 79-94.

[49]  J.E. Dickerson, J.A. Dickerson, "Fuzzy network profiling for intrusion detection," In Proc. of the 19th International Conference of the North American Fuzzy Information Processing Society (NAFIPS), Atlanta, GA, 2000, pp. 301-306.

[50]  M. Ramadas, S.O.B. Tjaden, "Detecting anomalous network traffic with self-organizing maps," In Proc. of the 6th International Symposium on Recent Advances in Intrusion Detection, Pittsburgh, PA, USA, 2003, pp. 36-54.

**Authors**

**Dewan Md. Farid** is currently a research fellow at ERIC Laboratory, University Lumière Lyon 2 - France. He obtained his B.Sc. Engineering in Computer Science and Engineering from Asian University of Bangladesh in 2003 and Master of Science in Computer Science and Engineering from United International University, Bangladesh in 2004. He is pursuing Ph.D. at Department of Computer Science and Engineering, Jahangirnagar University, Bangladesh. He is a faculty member in the Department of Computer Science and Engineering, United International University, Bangladesh. He has published nine international papers including one journal in the field of data mining, machine learning, and intrusion detection.

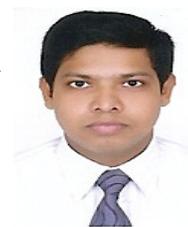

**Nouria Harbi** is a member of research staff ERIC Laboratory, University Lumière Lyon 2 - France. She received her master degree in computer science and Ph.D. from INSA Lyon - France. She then joined the laboratory ISEOR, where she worked on information systems. She is currently working on the security of decisional information system and modelling data warehouse.

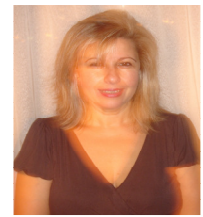

**Mohammad Zahidur Rahma** is currently a Professor at Department of Computer Science and Engineering, Jahangirnager University, Banglasesh. He obtained his B.Sc. Engineering in Electrical and Electronics from Bangladesh University of Engineering and Technology in 1986 and his M.Sc. Engineering in Computer Science and Engineering from the same institute in 1989. He obtained his Ph.D. degree in Computer Science and Information Technology from University of Malaya in 2001. He is a co-author of a book on E-commerce published from Malaysia. His current research includes the development of a secure distributed computing environment and e-commerce.

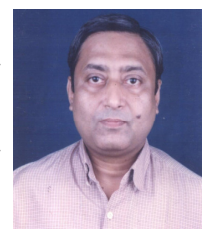